\documentclass{article}
\usepackage{spconf,amsmath,graphicx}

\usepackage{enumitem}
\setlist{nosep,leftmargin=14pt}
\usepackage{hyperref}       
\usepackage{url}            
\usepackage{booktabs}       
\usepackage{amsfonts}       
\usepackage{nicefrac}       
\usepackage{microtype}      
\usepackage{amsmath}
\usepackage{amssymb}
\usepackage{caption}
\usepackage{subcaption}
\usepackage[toc,page]{appendix}
\usepackage{floatrow}
\usepackage{pifont}
\newcommand{\cmark}{\ding{51}}%
\newcommand{\xmark}{\ding{55}}%

\usepackage{blindtext}
\usepackage{booktabs}
\usepackage{wrapfig,lipsum,booktabs}

\usepackage[ruled]{algorithm2e}

\usepackage{color}
\usepackage{multirow}
\usepackage{mwe} 

\title{Geometric Deep Learning on  Anatomical Meshes for the Prediction of Alzheimer's Disease}
\name{Ignacio Sarasua \thanks{First two authors contributed equally.} \qquad  Jonwong Lee \qquad Christian Wachinger}
\address{Lab for Artificial Intelligence in Medical Imaging (AI-Med), KJP, LMU M\"unchen, Germany }
\begin{document}
\maketitle
\begin{abstract}
    Geometric deep learning can find representations that are optimal for a given task and therefore improve the performance over pre-defined representations. 
    While current work has mainly focused on point representations, meshes also contain connectivity information and are therefore a more comprehensive characterization of the underlying anatomical surface. 
    In this work, we evaluate four recent geometric deep learning approaches that operate on mesh representations. 
    These approaches can be grouped into template-free and template-based approaches, where the template-based methods need a more elaborate pre-processing step with the definition of a common reference template and correspondences. 
    We compare the different networks for the prediction of  Alzheimer's disease based on the meshes of the hippocampus. 
    Our results show advantages for template-based methods in terms of accuracy, number of learnable parameters, and training speed. 
    While the template creation may be limiting for some applications, neuroimaging has a long history of building templates with automated tools readily available. 
    Overall, working with meshes is more involved than working with simplistic point clouds, but they also offer new avenues for designing geometric deep learning architectures.  

\end{abstract}

\begin{figure}
    \centering
    
    \includegraphics[width=\textwidth]{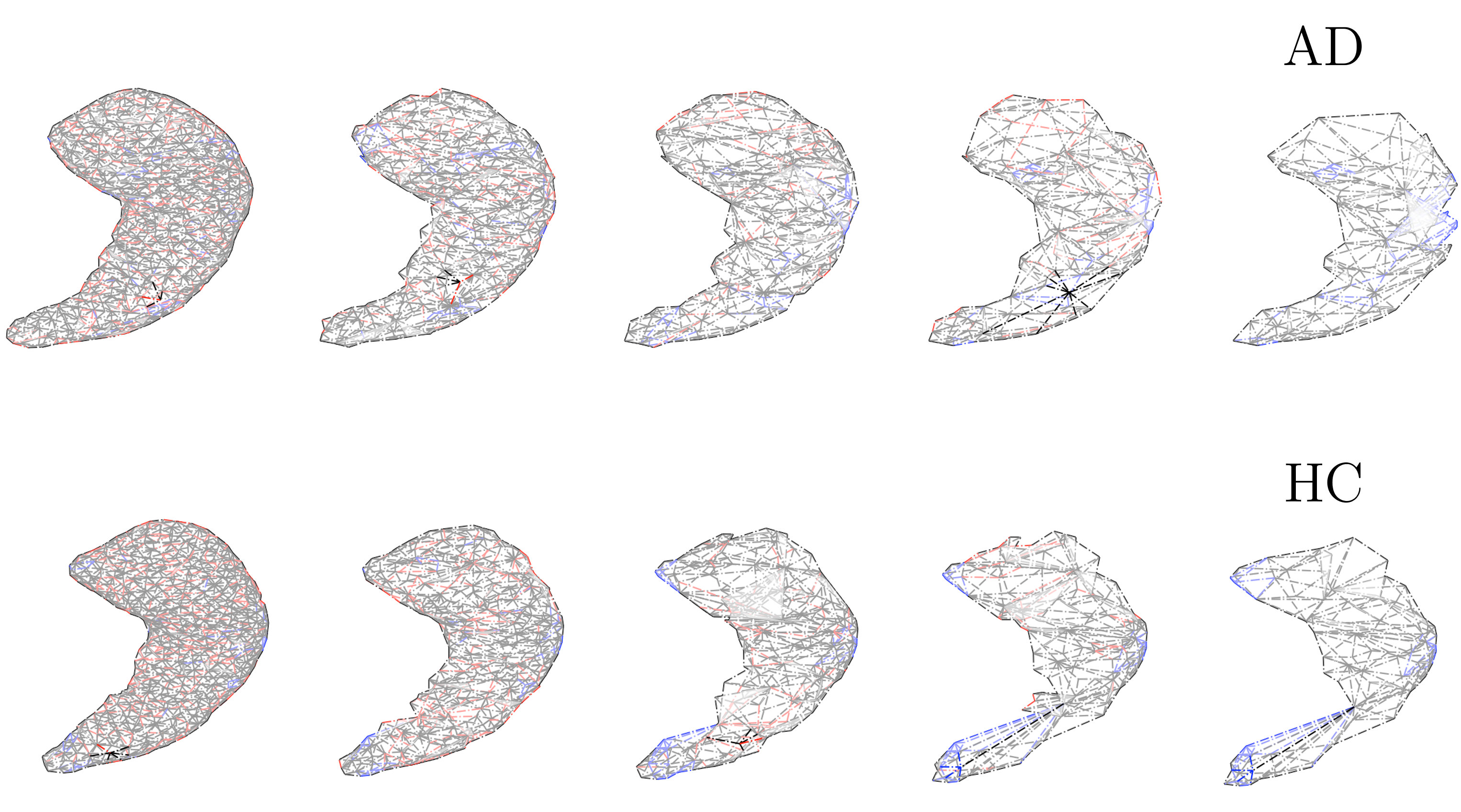}
    \caption{Difference between importance of edge features for an AD mesh (Top) and a HC mesh (bottom) in MeshCNN. Blue edges denote those with higher magnitude and red edges those with lower (and therefore pooled in the following step). }
    
    \label{fig:meshcnn_pool_hip}
\end{figure}
\section{Introduction}

Deep neural networks are offering new possibilities for shape analysis on medical data. 
They can learn a shape representation that is optimal for the given task, instead of relying on pre-defined shape representations~\cite{Bronstein2017}, and they scale to large datasets, enabling the identification of shape variations in wide populations.
The most common deep neural network for shape analysis is the PointNet~\cite{qi2017pointnet}, which takes point clouds as input. 
Recent results demonstrated that the PointNet can outperform traditional shape representations for disease prediction~\cite{gutierrez2018deep}. 

However, despite their simplicity, point cloud representations can be limited for capturing subtle shape changes or ambiguities. 
Hence, mesh representations have become more popular because they provide an efficient, non-uniform representation of a shape.
In contrast to point clouds, meshes have connectivity information, which establishes a more comprehensive representation of the underlying anatomical surface of the organ. 
Moreover, meshes can flexibly adapt to the complexity of the geometry. 
Salient  shape features that may be geometrically intricate can be captured by increasing the resolution, while large and simple regions can be represented by a small number of vertices and edges. 
Based on these  advantages of meshes in comparison to point clouds, a number of deep neural networks have been proposed in computer vision for learning on meshes~\cite{ranjan2018coma,gong2019spiralnet++,hanocka2019meshcnn,feng2019meshnet}. 

In this article, we want to study whether mesh representations also offer advantages over point clouds for disease prediction. 
In particular, we consider the \emph{template-based} methods CoMA~\cite{ranjan2018coma} and SpiralNet++~\cite{gong2019spiralnet++}, and the \emph{template-free} methods MeshCNN~\cite{hanocka2019meshcnn} and MeshNet~\cite{feng2019meshnet}.
To the best of our knowledge, this is the first application of these four methods on medical data. 
The template-based methods require a reference shape that shares the same structure with all the samples in the dataset, which is not required for the template-free approaches. 
We evaluate the different mesh-based models for the prediction of Alzheimer's disease and compare to the prediction based on point clouds. 
We also study the effect that having a fixed topology and reference template has on the performance of this task.
Our source code will be available after acceptance of the article\footnote{\url{https://github.com/ai-med/geomdl_anatomical_mesh}}.

\textbf{Related work}
Prior work in shape analysis for estimating discriminative models has mainly focused on the computation of handcrafted features~\cite{ng2014shape}, such as volume and thickness measures~\cite{costafreda2011automated}, medical descriptors~\cite{Gorczowski2007}, and spectral signatures~\cite{wachinger2016domain}. 
As an alternative, a variational auto-encoder was proposed to automatically extract features from 3D surfaces which were used for classification~\cite{Shakeri2016}.
Most related to our work are previous learning-based approaches that build on the PointNet~\cite{qi2017pointnet}. 
In~\cite{gutierrez2018deep,gutierrez2019learning},  deep neural networks have been introduced for classifying between healthy controls and AD patients using anatomical point clouds as input.



\label{sec:intro}


\section{Methods} 
\label{sec:method}
A 3D mesh, $M = (V,E,A,F)$, is defined by a set of vertices, $V\in\mathbb{R}^{N\times 3}$, and edges, $E$, that connect the vertices. $A\in\{0,1\}^{N \times N}$ is the adjacency matrix that indicates the connection between two vertices (e.g. $A_{i,j}=1$ if vertex $i$ is connected to vertex $j$ and $0$ otherwise) and $F$ are the faces formed by a set of edges (three in case of triangular meshes). As opposed to images, meshes are not represented on a regular grid of discrete values and therefore common CNN operations such as \emph{convolution} and \emph{pooling}  are not explicitly defined anymore. In particular, defining local neighborhood for a given vertex becomes challenging. In this section, we introduce four methods that overcome this issue by re-formalizing these operations for triangular meshes. We divide the approaches according to their need to register to a template mesh. We summarize the main contributions of each method, without going into much detail due to space constrains. For a deeper understanding of each approach, as well as implementation details, we encourage the reader to peruse the original authors' work. 

\subsection{Template-based methods}
Template-based models require correspondences to a  reference shape that shares the same structure with all the samples in the dataset. This means that all  mesh samples have the same number of vertices and the same connectivity between them. 
An advantage of template-based models is that operations can be pre-computed on the template and afterwards applied to the individual samples. For example, both methods described in this section define the down-sampling operation by \emph{Minimum Quadric Error Pooling}, where vertex pairs are iteratively contracted, such that would minimize the quadric error \cite{garland1997surface}. For efficiency, the coordinates of the vertices that must be pooled in each level are computed for the template and then applied to the samples in the dataset.
This  significantly reduces both the training and inference time, as well as the complexity of the model. 

\textbf{Convolutional Mesh Encoder}\label{sec:coma}
As first model, we built on the Convolutional Mesh Autoncoder (CoMA)~\cite{ranjan2018coma}. Even though this architecture was originally proposed for generative tasks, we will use the encoder path of their architecture (we will refer to this modified version as Convolutional Mesh Encoder or CoME) and add a multi-layer-perceptron (MLP) to transform the latent features into classification labels. In order to overcome the issue of defining a local neighborhood, CoME deploys \emph{fast spectral convolutions} \cite{defferrard2016convolutional}, which operate in the frequency domain. Fast spectral convolutions use Chebyshev polynomials \cite{hammond2011wavelets} for parametrizing the convolutional filters to reduce the operation's complexity.

\textbf{SpiralNet++}
Gong et al.~\cite{gong2019spiralnet++}  proposed SpiralNet++ that has a novel \emph{message passing} approach to deal with irregular representations, such as meshes.  In contrast to CoME, SpiralNet++ stays in the spatial domain, where it defines \emph{spiral sequences} within the vertex neighborhood. These are pre-computed only once in the template and applied to the rest of the dataset. The\emph{spiral convolutional operator} is defined by concatenating the features of the vertices within the spiral sequence, and passing them through a MLP.

\subsection{Template-free methods}
Despite the advantages of template based models, defining a common topology for the whole dataset can become cumbersome in certain instances. Non-template based models do not require a reference shape that shares the structure with all the other samples. In this section, we introduce two approaches that explore other characteristics of mesh representations without having to define a fixed topology.

\textbf{MeshCNN}
MeshCNN~\cite{hanocka2019meshcnn} operates directly on the mesh edges, $E$, by defining a set of the features associated to them.  Edge features provide non-uniform, geodesic neighborhood information and a consistent number of neighbors. In the interest of solving order ambiguity, the input descriptors of each edge are designed such that they only contain relative geometric features, and aggregated in pairs applying simple symmetric operations (e.g. addition). The convolution operation is computed by using invariant kernels and those edges with less relative influence (smallest magnitude of edge features) are collapsed during the pooling step.

\textbf{MeshNet}
MeshNet~\cite{feng2019meshnet} is a face-based approach. The main idea is to consider the faces of the mesh as the main feature unit, which provides advantages over the other methods in terms of regularity of the neighborhood and ordering invariance. MeshNet defines two types of features: \emph{spatial} and \emph{structural}.  Spatial features are computed by passing the coordinates of the face's center through a shared MLP. Structural features focus  on  obtaining  ’inner’  and  ’outer’  information  from the  face.  Inner  information  is  obtained  by  an  operation  defined  as \emph{face rotate convolution}. The 'outer' information of the face's structure is obtained by \emph{face kernel correlation}. The \emph{mesh convolution} is conducted by combination and aggregation operations. The combination operation concatenates the outputs from spatial and structural descriptors and passes them through a shared MLP. In the aggregation block, the structural features from the target face and its neighbors are collected and concatenated.  An additional shared MLP is applied to each concatenated vector. Then, the outputs of the shared MLP are aggregated by max pooling. After the aggregation operation for each face, a shared MLP is applied and provides the final output of the mesh convolution.

\section{Experiments}
We evaluate the performance of the methods in distinguishing healthy controls (HC) from patients diagnosed with Alzheimer's diseases (AD).

\textbf{Data}
For all our experiments, we use the shape of the left hippocampus, given its  importance in AD pathology~\cite{thompson2004mapping}. To obtain the hippocampi point clouds, we segment image data from the Alzheimer’s Disease Neuroimaging Initiative (ADNI) database (adni.loni.usc.edu)~\cite{Jack2008} using FIRST \cite{patenaude2007bayesian} from  the Functional Magnetic Resonance Imaging of the Brain, Software Library FSL, which also provides the meshes for the segmented samples. FIRST segments the subcortical structures by registering them to a reference template, creating voxel-wise correspondences (and therefore, also vertex-wise) between the template and every sample in the dataset. 
This is particularly useful for template-based methods like CoME or SpiralNet++, since it forces the topology of all the samples to be the same and provides us with a template shape. The dataset of 282 AD patients and 282 healthy controls was divided into training and testing sets (50/50 split).

\textbf{Results} In Table \ref{tab:results}, we present the accuracy, precision and recall of each method for AD classification, as well as the number of trainable parameters. We compare the methods to PointNet \cite{qi2017pointnet}, since it is the state-of-the-art for shape-based AD classification  on the  ADNI dataset \cite{gutierrez2018deep}.
The training time per epoch for the methods is about 2 minutes for MeshCNN,  5 seconds for MeshNet,  3 seconds for CoMe,  0.1 seconds for SpiralNet++, and 1 second for PointNet. 

We observe that the template-based methods have a better performance than the template-free methods and PointNet. 
Moreover, the pre-registration with a template, make SpiralNet++ and CoME much more efficient in terms of complexity. Their pre-definition of the neighborhood on the template, allows these techniques to define convolutional operations very similar to regular CNN. 
Therefore, they do not require a fully-connected layers. We also observe such phenomena when we compare MeshCNN against MeshNet, where the former, while having more learnable parameters than the template-based techniques due to the depth of the filters, require a much smaller number of parameters than the latter.
SpiralNet++ has the overall highest accuracy and slightly outperforms CoME. 
MeshNet has a similar accuracy than PointNet and both of them outperform MeshCNN. 
Overall, template-based methods have the highest accuracy, lowest number of parameters, and the fastest training time. 
PointNet, which also does not use a template, has a similar accuracy to the template-free MeshNet.



We would also like to emphasize that, while MeshCNN reports the lowest performance for this task, its pooling operation allows us to visualize which areas of the shape are more and less relevant to drive a decision. In Figure \ref{fig:meshcnn_pool_hip}, we show the evolution throughout the network for AD and HC hippocampi. We  can observe that the edges on the medial part of the body in the subiculum area, the lateral part of the body in the CA1 area and the inferior part of the hippocampus head in the subiculum area are assigned higher weights, which is consistent with prior results on AD related changes \cite{thompson2004mapping}.
\begin{table}
\small
\caption{Classification results (accuracy, precision, recall) for PointNet and the four mesh-based neural networks (template-based and template-free) with the number of parameters.} \label{tab:results}
\centering
\begin{tabular}{l@{\hskip 0.05in}@{\hskip 0.05in}c@{\hskip 0.05in}c@{\hskip 0.05in}c@{\hskip 0.05in}c@{\hskip 0.05in}c@{\hskip 0.05in}r}
\toprule
Method     & Data       & Template           & Acc(\%)               & Prec      & Rec       & \# Params.     \\ 
\midrule
PointNet    &         &\xmark            & $73.5$             & $0.74$        & $0.72$            &$1600K$    \\ 
MeshCNN     & Mesh     &\xmark             & $71.3$            & $\mathbf{0.82}$        & $0.66$            &$165K$  \\ 
MeshNet     & Mesh     &\xmark             & $73.8$            & $\mathbf{0.82}$        & $0.66$            &$4250K$    \\ 
CoME        & Mesh     &\cmark             & $77.0$            & $0.75$        & $\mathbf{0.80}$            &$\mathbf{44K}$   \\
SpiralNet++ & Mesh     &\cmark             & $\mathbf{77.3}$            & $0.78$        & $0.77$            &$85K$   \\
\bottomrule
\end{tabular}
\end{table} 
\section{Conclusion}
In this work, we have compared four state-of-the-art methods for mesh-based analysis of anatomical shapes. 
Most of the mesh-based methods need a much smaller number of learnable parameters than the point-based PointNet. 
In particular, the template-based CoME and SpiralNet++ have much fewer parameters. 
This is mainly due to the definition of convolution layers on meshes, which are more compact than fully connected layers. 
The template-based methods also have the highest accuracy and the fastest training time. 
However, template-based methods need a more elaborate pre-processing stage with the definition of a template and the computation of correspondences. 
Nevertheless, medical imaging and in particular neuroimaging has a long history in the creation of templates and the variation in the population is commonly not as dramatic as in computer vision applications, so that the restrictions of template-based methods do not weigh too heavily. 
The field of geometric deep learning is rapidly evolving so that future mesh-based networks may further push current performance boundaries. 

\section{Acknowledgements}
This research was supported by BMBF and the Bavarian State Ministry of Science and the Arts and coordinated by the Bavarian Research Institute for Digital Transformation (bidt). The authors have no relevant financial or non-financial interests to disclose

\section{Compliance with Ethical Standards}
This research study was conducted retrospectively using human subject data made available in open access by \cite{Jack2008}. Ethical approval was not required as confirmed by the license attached with the open access data.

\bibliographystyle{IEEEbib}
\bibliography{strings.bib}

\end{document}